\documentclass[11pt,a4paper]{article}
\usepackage[hyperref]{emnlp2020}
\usepackage{times}
\usepackage{latexsym}

\usepackage{microtype}

\usepackage{adjustbox}
\usepackage{url}
\usepackage[ruled]{algorithm2e}
\usepackage{amssymb}
\usepackage{amsmath}
\usepackage{arydshln}
\usepackage{multirow}
\usepackage{graphicx} 
\graphicspath{ {./images/} }

\newcommand{\Hmat}{{\bf H}}

\newcommand{\Mmat}{{\bf M}}

\newcommand{\Xmat}{{\bf X}}

\newcommand{\hv}{{\boldsymbol h}}

\newcommand{\mv}{{\boldsymbol m}}

\newcommand{\xv}{{\boldsymbol x}}

\newcommand{\zv}{{\boldsymbol z}}

\newcommand{\thetav}{{\boldsymbol \theta}}

\newcommand{\R}{\mathbb{R}}

\newcommand{\Lcal}{\mathcal{L}}

\definecolor{amber}{rgb}{1.0, 0.49, 0.0}
\definecolor{applegreen}{rgb}{0.55, 0.71, 0.0}
\aclfinalcopy 
\def\aclpaperid{3394} 

\title{Contrastive Distillation on Intermediate Representations \\ for Language Model Compression}

\author{\textbf{Siqi Sun}, \hspace{1mm} \textbf{Zhe Gan}, \hspace{1mm} \textbf{Yu Cheng},\hspace{1mm} \textbf{Yuwei Fang},\hspace{1mm} \textbf{Shuohang Wang}, \hspace{1mm} \textbf{Jingjing Liu}\\
Microsoft Dynamics 365 AI Research\\
{\tt \small{ \{siqi.sun,zhe.gan,yu.cheng,yuwfan,shuohang.wang,jingjl\}@microsoft.com}}}

\date{}

\begin{document}
\maketitle

\begin{abstract}
Existing language model compression methods mostly use a simple $L_2$ loss to distill knowledge in the intermediate representations of a large BERT model to a smaller one. Although widely used, this objective by design assumes that all the dimensions of hidden representations are independent, failing to capture important \emph{structural} knowledge in the intermediate layers of the teacher network. To achieve better distillation efficacy, we propose Contrastive Distillation on Intermediate Representations (\textsc{CoDIR}), a principled knowledge distillation framework where the student is trained  to distill knowledge through intermediate layers of the teacher via a \emph{contrastive objective}. By learning to distinguish positive sample from a large set of negative samples,
CoDIR facilitates the student's exploitation of rich information in teacher's hidden layers. CoDIR can be readily applied to compress large-scale language models in both pre-training and finetuning stages, and achieves superb performance on the GLUE benchmark, outperforming state-of-the-art compression methods.\footnote{Code will be released at \url{https://github.com/intersun/CoDIR}.}
\end{abstract}

\section{Introduction}
Large-scale pre-trained language models (LMs), such as BERT~\cite{devlin2018bert}, XLNet~\cite{yang2019xlnet} and RoBERTa~\cite{liu2019roberta}, have brought revolutionary advancement to the NLP field~\cite{wang2018glue}. However, as new-generation LMs grow more and more into behemoth size, it becomes increasingly challenging to deploy them in resource-deprived environment. Naturally, there has been a surge of research interest in developing model compression methods~\cite{sun2019patient,sanh2019distilbert,shen2019q} to reduce network size in pre-trained LMs, while retaining comparable performance and efficiency.  

PKD~\citep{sun2019patient} was the first known effort in this expedition, an elegant and effective method that uses knowledge distillation (KD) for BERT model compression at finetuning stage. 
Later on, DistilBERT~\cite{sanh2019distilbert}, TinyBERT~\cite{jiao2019tinybert} and MobileBERT~\cite{sun2020mobilebert} carried on the torch and extended similar compression techniques to pre-training stage, allowing efficient training of task-agnostic compressed models. 
In addition to the conventional KL-divergence loss applied to the probabilistic output of the teacher and student networks, an
$L_2$ loss measuring the difference between normalized hidden layers has proven to be highly effective in these methods. 
However, $L_2$ norm follows the assumption that all dimensions of the target representation are independent, which overlooks important structural information in the many hidden layers of BERT teacher. 

Motivated by this, we propose \textbf{Co}ntrastive \textbf{D}istillation for \textbf{I}ntermediate \textbf{R}epresentations (\textsc{CoDIR}), which uses a \emph{contrastive objective} to capture higher-order output dependencies between intermediate representations of BERT teacher and the student. 
Contrastive learning~\cite{gutmann2010noise} aims to learn representations by enforcing similar elements to be equal and dissimilar elements further apart. Formulated in either supervised or unsupervised way, it has been successfully applied to diverse applications \cite{hjelm2018learning,he2019momentum,tian2019contrastive,khosla2020supervised}. To the best of our knowledge, utilizing contrastive learning to compress large Transformer models is still an unexplored territory, which is the main focus of this paper.

A teacher network's hidden layers usually contain rich semantic and syntactic knowledge that can be instrumental if successfully passed on to the student~\cite{tenney2019bert,kovaleva2019revealing,sun2019patient}. Thus, instead of directly applying contrastive loss to the final output layer of the teacher, we apply contrastive learning to its intermediate layers, in addition to the use of KL-loss between the probabilistic outputs of the teacher and student. This casts a stronger regularization effect for student training by capturing more informative signals from intermediate representations. To maximize the exploitation of intermediate layers of the teacher, we also propose the use of mean-pooled representation as the distillation target, which is empirically more effective than commonly used special \texttt{[CLS]} token. 

\begin{figure*}[t!]
\centering
{\includegraphics[width=\linewidth]{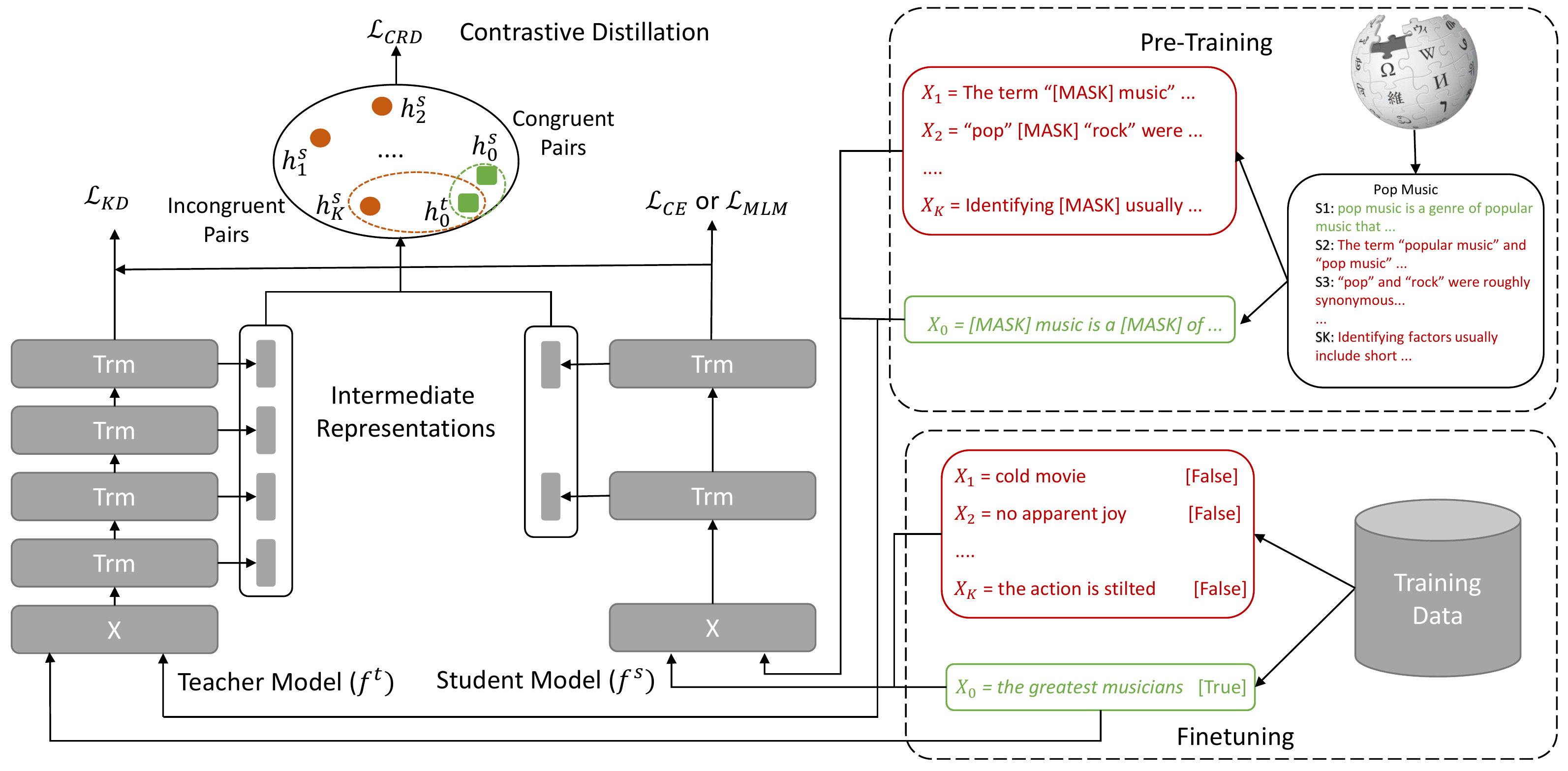}}
\caption{\label{fig:codir}Overview of the proposed CoDIR framework for language model compression in both pre-training and finetuning stages. ``Trm'' represents a Transformer block, $X$ are input tokens, $f^t$ and $f^s$ are teacher and student models, and $X_0, \{X_i\}_{i=1}^K$ represent one positive sample and a set of negative samples, respectively. The difference between CoDIR pre-training and finetuning mainly lies in the negative example sampling method.}
\end{figure*}

To realize constrastive distillation, we define a \emph{congruent} pair ($\hv_{i}^t,\hv_{i}^s$) as the pair of representations of the same data input 
from the teacher and student networks, as illustrated in Figure~\ref{fig:codir}. \emph{Incongruent} pair ($\hv_{i}^t,\hv_{j}^s$) is defined as the pair of representations of two different data samples through the teacher and the student networks, respectively. 
The goal is to train the student network to distinguish the congruent pair from a large set of incongruent pairs, by minimizing the constrastive objective. 

For efficient training, all data samples are stored in a memory bank~\cite{wu2018unsupervised,he2019momentum}. 
During finetuning, incongruent pairs can be selected by choosing sample pairs with different labels to maximize the distance.
For pre-training, however, it is not straightforward to construct incongruent pairs this way as labels are unavailable. Thus, we randomly sample data points from the same mini-batch pair to form incongruent pairs, and construct a proxy congruent-incongruent sample pool to assimilate what is observed in the downstream tasks during finetuning stage. 
This and other designs in CoDIR make constrative learning possible for LM compression, and have demonstrated strong performance and high efficiency in experiments.

Our contributions are summarized as follows. ($i$) We propose \textsc{CoDIR}, a principled framework to distill knowledge in the intermediate representations of large-scale language models via a contrastive objective, instead of a conventional $L_2$ loss. ($ii$) We propose effective sampling strategies to enable CoDIR in both pre-training and finetuning stages. ($iii$) Experimental results demonstrate that CoDIR can successfully train a half-size Transformer model that achieves competing performance to BERT-base on the GLUE benchmark~\cite{wang2018glue}, with half training time and GPU demand. Our pre-trained model checkpoint will be released for public access. 

\section{Related Work}
\paragraph{Language Model Compression} To reduce computational cost of training large-scale language models, many model compression techniques have been developed, such as quantization \cite{shen2019q, zafrir2019q8bert}, pruning \cite{guo2019reweighted, gordon2020compressing, michel2019sixteen}, knowledge distillation \cite{tang2019distilling, sun2019patient, sanh2019distilbert, jiao2019tinybert, sun2020mobilebert}, and direct Transformer block modification \cite{kitaev2020reformer, wu2020lite}. 

Quantization refers to storing model parameters from 32- or 16-bit floating number to 8-bit or even lower. Directly truncating the parameter values will cause significant accuracy loss, hence quantization-aware training has been developed to maintain similar accuracy to the original model \cite{shen2019q, zafrir2019q8bert}. \citet{michel2019sixteen} found that even after most attention heads are removed, the model still retains similar accuracy, indicating there is high redundancy in the learned model weights. Later studies proposed different pruning-based methods. For example, \citet{gordon2020compressing} simply removed the model weights that are close to zero; while \citet{guo2019reweighted} used re-weighted $L_1$ and proximal algorithm to prune weights to zero. Note that simple pruning does not improve inference speed, unless there is structure change such as removing the whole attention head. 

There are also some efforts that try to improve the Transformer block directly. Typically, language models such as BERT~\cite{devlin2018bert} and RoBERTa~\cite{liu2019roberta} can only handle a sequence of tokens in length up to 512. \citet{kitaev2020reformer} proposed to use reversible residual layers and locality-sensitive hashing to reduce memory usage to deal with extremely long sequences. Besides, \citet{wu2020lite} proposed to use convolutional neural networks to capture short-range attention such that reducing the size of self-attention will not significantly hurt performance.

Another line of research on model compression is based on knowledge transfer, or knowledge distillation (KD) \cite{hinton2015distilling}, which is the main focus of this paper. Note that previously introduced model compression techniques are orthogonal to KD, and can be bundled for further speedup. Distilled BiLSTM~\cite{tang2019distilling} tried to distill knowledge from BERT into a simple LSTM. Though achieving more than 400 times speedup compared to BERT-large, it suffers from significant performance loss due to the shallow network architecture. DistilBERT \cite{sanh2019distilbert} proposed to distill predicted logits from the teacher model into a student model with 6 Transformer blocks. BERT-PKD \cite{sun2019patient} proposed to not only distill the logits, but also the representation of \texttt{[CLS]} tokens from the intermediate layers of the teacher model. TinyBERT \cite{jiao2019tinybert}, MobileBERT \cite{sun2020mobilebert} and SID \cite{aguilar2019knowledge} further proposed to improve BERT-PKD by distilling more internal representations to the student, such as embedding layers and attention weights. These existing methods can be generally divided into two categories: ($i$) task-specific, and ($ii$) task-agnostic. Task-specific methods, such as Distilled BiLSTM, BERT-PKD and SID, require the training of individual teacher model for each downstream task; while task-agnostic methods such as DistilBERT, TinyBERT and MobileBERT use KD to pre-train a model that can be applied to all downstream tasks by standard finetuning. 

\paragraph{Contrastive Representation Learning} 
Contrastive learning~\cite{gutmann2010noise,arora2019theoretical} is a popular research area that has been successfully applied to density estimation and representation learning, especially in self-supervised setting~\cite{he2019momentum,chen2020simple}. It has been shown that the contrastive objective can be interpreted as maximizing the lower bound of mutual information between different views of the data~\cite{hjelm2018learning,oord2018representation,bachman2019learning,henaff2019data}. However, it is unclear whether the success is determined by mutual information or by the specific form of the contrastive loss~\cite{tschannen2019mutual}.  Recently, it has been extended to knowledge distillation and cross-modal transfer for image classification tasks~\cite{tian2019contrastive}. Different from prior work, we propose the use of contrastive objective for Transformer-based model compression and focus on language understanding tasks. 


\section{CoDIR for Model Compression}
In this section, we first provide an overview of the proposed method in Sec.~\ref{sec:overview}, then describe the details of contrastive distillation in Sec.~\ref{sec:distil_obj}. Its adaptation to pre-training and finetuning is further discussed in Sec.~\ref{sec:pretrain_and_finetune}.

\subsection{Framework Overview} \label{sec:overview}
We use RoBERTa-base~\cite{liu2019roberta} as the teacher network, denoted as $f^t$, which has 12 layers with 768-dimension hidden representations. We aim to transfer the knowledge of $f^t$ into a student network $f^s$, where $f^s$ is a 6-layer Transformer~\cite{vaswani2017attention} to mimic the behavior of $f^t$~\cite{hinton2015distilling}.
Denote a training sample as $(X,y)$, where $X= (x_0,\ldots,x_{L-1})$ is a sequence of tokens in length $L$, and $y$ is the corresponding label (if available). The word embedding matrix of $X$ is represented as $\Xmat=(\xv_{0},\ldots,\xv_{L-1})$, where $\xv_{i} \in \R^d$ is a $d$-dimensional vector, and $\Xmat \in \R^{L\times d}$. In addition, the intermediate representations at each layer for the teacher and student are denoted as $\Hmat^t=(\Hmat^t_1, \dots, \Hmat^t_{12})$ and $\Hmat^s=(\Hmat^s_1, \dots, \Hmat^s_6)$, respectively, where $\Hmat_i^t, \Hmat_i^s \in \mathbb{R}^{L \times d}$ contains all the hidden states in one layer. And $\zv^t, \zv^s\in\R^k$ are the logit representations (before the softmax layer) of the teacher and student, respectively, where $k$ is the number of classes. 


As illustrated in Figure~\ref{fig:codir}, our distillation objective consists of three components: ($i$) original \emph{training loss} from the target task; ($ii$) conventional \emph{KL-divergence-based loss} to distill the knowledge of $\zv^t$ into $\zv^s$; ($iii$) proposed \emph{contrastive loss} to distill the knowledge of $\Hmat^t$ into $\Hmat^s$. The final training objective can be written as: 
\begin{align} \label{eqn:overall_objective}
    \Lcal_{\text{CoDIR}}(\thetav) &= \Lcal_{\text{CE}} (\zv^s, y;\thetav) + \alpha_1 \Lcal_{\text{KD}} (\zv^t,\zv^s;\thetav) \nonumber \\
    &+ \alpha_2 \Lcal_{\text{CRD}} (\Hmat^t, \Hmat^s; \thetav)\,, 
\end{align}
where $\Lcal_{\text{CE}}, \Lcal_{\text{KD}}$ and $\Lcal_{\text{CRD}}$ correspond to the original loss, KD loss and contrastive loss, respectively. $\thetav$ denotes all the learnable parameters in the student $f^s$, while the teacher network is pre-trained and kept fixed. $\alpha_1, \alpha_2$ are two hyper-parameters to balance the loss terms.

$\Lcal_{\text{CE}}$ is typically implemented as a cross-entropy loss for classification problems, and $\Lcal_{\text{KD}}$ can be written as 
\begin{align}
    \Lcal_{\text{KD}} (\zv^t,\zv^s;\thetav) = \mathrm{KL}\left( g(\zv^t / \rho) \| g(\zv^s / \rho)  \right)\,, 
\end{align}
where $g(\cdot)$ denotes the softmax function, and $\rho$ is the temperature. $\Lcal_{\text{KD}}$ encourages the student network to produce distributionally-similar outputs to the teacher network. 

Only relying on the final logit output for distillation discards the rich information hidden in the intermediate layers of BERT. 
Recent work~\cite{sun2019patient,jiao2019tinybert} has found that distilling the knowledge from intermediate representations with $L_2$ loss can further enhance the performance. Following the same intuition, our proposed method also aims to achieve this goal, with a more principled contrastive objective as  detailed below. 

\subsection{Contrastive Distillation} \label{sec:distil_obj}

First, we describe how to summarize intermediate representations into a concise feature vector. Based on this, we detail how to perform contrastive distillation~\cite{tian2019contrastive} for model compression.

\paragraph{Intermediate Representation}
Directly using $\Hmat^t$ and $\Hmat^s$ for distillation is infeasible, as the total feature dimension  is $|\Hmat^s|=6 \times 512 \times 768\approx 2.4$ million for a sentence in full length (\emph{i.e.}, $L=512$). Therefore, we propose to first perform mean-pooling over $\Hmat^t$ and $\Hmat^s$ to obtain a layer-wise sentence embedding. Note that the embedding of the \texttt{[CLS]} token can also be used directly for this purpose; however, in practice we found that mean-pooling performs better.
Specifically, 
we conduct row-wise average over $\Hmat^t_i$ and $\Hmat^s_i$: 
\begin{align}
    \bar{\hv}_i^t = \text{Pool}(\Hmat_i^t), \,\,
    \bar{\hv}_i^s = \text{Pool}(\Hmat_i^s)\,,
\end{align}
where $\bar{\hv}_i^t, \bar{\hv}_i^s $ $\in \mathbb{R}^d$ are the sentence embedding for layer $i$ of the teacher and student model, respectively. Therefore, the student's intermediate representation can be summarized as $\bar{\hv}^s=[\bar{\hv}_1^s; \dots; \bar{\hv}_6^s] \in \mathbb{R}^{6d}$, where $[;]$ denotes vector concatenation. Similarly, the teacher's intermediate representation can be summarized as $\bar{\hv}^t=[\bar{\hv}_1^t; \dots; \bar{\hv}_{12}^t] \in \mathbb{R}^{12d}$. Two linear mappings $\phi^s:\mathbb{R}^{6d} \rightarrow \mathbb{R}^{m}$ and $\phi^t: \mathbb{R}^{12d} \rightarrow \mathbb{R}^{m}$ are further applied to project $\bar{\hv}^t$ and $\bar{\hv}^s$ into the same low-dimensional space, yielding $\hv^t, \hv^s \in \R^{m}$, which are used for calculating the contrastive loss.

\paragraph{Contrastive Objective}
Given a training sample $(X_0, y_0)$, we first randomly select $K$ negative samples with different labels, denoted as $\{(X_i,y_i)\}_{i=1}^K$.  
Following the above process, we can obtain a summarized intermediate representation $\hv_0^t, \hv_0^s \in \R^{m}$ by sending $X_0$ to both the teacher and student network. Similarly, for negative samples, we can obtain $\{\hv_i^s\}_{i=1}^K$.   

Contrastive learning aims to map the student's representation $\hv_0^s$ close to the teacher's representation $\hv_0^t$, while the negative samples' representations $\{\hv_i^s\}_{i=1}^K$ far apart from $\hv_0^t$. 
To achieve this, we use the following InfoNCE loss~\cite{oord2018representation} for model training:
\begin{align}
\mathcal{L}_{\text{CRD}} (\thetav) = - \log \frac{\exp{\big(\langle\hv_0^t, \hv_0^s\rangle /\tau}\big)}{\sum_{j=0}^K \exp{\big(\langle\hv_0^t, \hv_j^s\rangle /\tau}\big)}\,,
\end{align}
where $\langle \cdot,\cdot\rangle$ denotes the cosine similarity between two feature vectors, and $\tau$ is the temperature that controls the concentration level. As demonstrated, contrastive distillation is implemented as a $(K+1)$-way classification task, which is interpreted as maximizing the lower bound of mutual information between $\hv_0^t$ and $\hv_0^s$~\cite{oord2018representation,tian2019contrastive}.



\subsection{Pre-training and Finetuning Adaptation} \label{sec:pretrain_and_finetune}

\paragraph{Memory Bank}
For a positive pair $(\hv_0^t, \hv_0^s)$, one needs to compute the intermediate representations for all the negative samples, \emph{i.e.}, $\{\hv_i^s\}_{i=1}^K$, which requires $K+1$ times computation compared to normal training. A large number of negative samples is required to ensure performance~\cite{arora2019theoretical}, which renders large-scale contrastive distillation infeasible for practical use. 
To address this issue, we follow~\citet{wu2018unsupervised} and use a memory bank $\Mmat \in \mathbb{R}^{N \times m}$ to store the intermediate representation of all $N$ training examples, and the representation is only updated for positive samples in each forward propagation. Therefore, the training cost is roughly the same as in normal training. Specifically, assume the mini-batch size is 1, then at each training step, $\Mmat$ is updated as: 
\begin{align}
    \mv_{0} = \beta \cdot \mv_{0} + (1-\beta) \cdot \hv_0^s\,,
\end{align}
where $\mv_0$ is the retrieved representation from memory bank $\Mmat$ that corresponds to $\hv_0^s$, and $\beta \in (0,1)$ is a hyper-parameter that controls how aggressively the memory bank is updated.


\paragraph{Finetuning}
Since task-specific label supervision is available in finetuning stage, applying CoDIR to finetuning is relatively straightforward. When selecting negative samples from the memory bank, we make sure the selected samples have different labels from the positive sample. 

\paragraph{Pre-training}
For pre-training, the target task becomes masked language modeling (MLM)~\cite{devlin2018bert}. Therefore, we replace the $\Lcal_{\text{CE}}$ loss in Eqn.~(\ref{eqn:overall_objective}) with $\Lcal_{\text{MLM}}$.
Following \citet{liu2019roberta,lan2019albert}, we did not include the next-sentence-prediction task for pre-training, as it does not improve performance on downstream tasks. 
Since task-specific label supervision is unavailable during pre-training, we propose an effective method to select negative samples from the memory bank. Specifically, we sample negative examples randomly from the same mini-batch each time, as they have closer semantic meaning as some of them are from the same article, especially for Bookcorpus \cite{zhu2015aligning}. Then, we use the sampled negative examples to retrieve representations from the memory bank. Intuitively, negative examples sampled in this way serve as ``hard'' negatives, compared to randomly sampling from the whole training corpus; otherwise, the $\Lcal_{\text{CRD}}$ loss could easily drop to zero if the task is too easy.


\begin{table*}[t!]
\centering
\resizebox{1.0\textwidth}{!}{
\begin{tabular}{|c c c c c c c c c|} 
\hline
 \multirow{2}{*}{Model} &        CoLA & SST-2 & MRPC & QQP & MNLI-m/-mm & QNLI & RTE &  \multirow{2}{*}{Ave.} \\ 
 [0.5pt] & (8.5k) & (67k) & (3.7k) & (364k) & (393k) & (108k) & (2.5k) & \\
\hline\hline
RoBERTa-base (Ours) & 60.3 & 95.3 & 91.0 & 89.6 & 87.7/86.8 & 93.5 & 71.7 & 84.5\\
BERT-base (Google)     & 52.1	& 93.5	& 88.9	& 89.2    & 84.6/83.4  & 90.5  & 66.4	& 81.7 \\
 \hline
 DistilBERT   & 32.8	& 91.4	& 82.4	& 88.5   & 78.9/78.0  & 85.2	& 54.1	& 73.9 \\
 SID          & 41.4    & -     & 83.8    & 89.1   & -     &   -   & 62.2  &  - \\
 BERT$_6$-PKD & 24.8    & 92.0	& 86.4	& 88.9   & 81.5/81.0	& 89.0	& 65.5  & 76.0\\	 

 TinyBERT$_4^\star$ & 43.3    & 92.6	& 86.4	& \textbf{89.2$^*$}	    & 82.5/81.8	& 87.7	& 62.9	& 76.7 \\
 TinyBERT$_6$ & 51.1$^*$      & 93.1  & 87.3    & 89.1    & \textbf{84.6/83.2}  & \textbf{90.4}  & 66.0  & 80.6\\
 \hline
 MLM-Pre + Fine           & 50.6    & 93.0	& 88.7  & \textbf{89.2}	& 82.9/82.0	& 89.6	& 62.1	& 79.8 \\
 CoDIR-Fine          & 53.6	& 93.6	& 89.4	& 89.1	& 83.6/82.8	& \textbf{90.4}	& 65.6	& 81.0 \\
 CoDIR-Pre           & \textbf{53.7}	& \textbf{94.1}	& 89.3	& 89.1	& 83.7/82.6	& \textbf{90.4}	& 66.8	& \textbf{81.2} \\

 CoDIR-Pre + CoDIR-Fine  &  \textbf{53.7} & 93.6  &  \textbf{89.6} & 89.1 & 83.5/82.7 & 90.1 & \textbf{67.1} & \textbf{81.2} \\
 \hline
\end{tabular}
}
\caption{Results on GLUE Benchmark. (*) indicates those numbers are unavailable in the original papers and were obtained by us through submission to the official leaderboard using their codebases. Other results are obtained from published papers. ($\star$) indicates those methods with fewer Transformer blocks, and may not be fair comparison.} 
\label{tab_glue_test}
\end{table*}

\section{Experiments}
In this section, we present comprehensive experiments on a wide range of downstream tasks and provide detailed ablation studies, to demonstrate the effectiveness of the proposed approach to large-scale LM compression.

\subsection{Datasets}
We evaluate the proposed approach on sentence classification tasks from the General Language Understanding Evaluate (GLUE) benchmark \cite{wang2018glue}, as our finetuning framework is designed for classification, and we only exclude the STS-B dataset \cite{cer2017semeval}. Following other works \cite{sun2019patient, jiao2019tinybert, sun2020mobilebert}, we also do not run experiments on WNLI dataset \cite{levesque2012winograd}, as it is very difficult and even majority voting outperforms most benchmarks.\footnote{Please refer to https://gluebenchmark.com/leaderboard.} 

\paragraph{CoLA} Corpus of Linguistic Acceptability \cite{warstadt2019neural} contains a collection of 8.5k sentences drawn from books or journal articles. The goal is to predict if the given sequence of words is grammatically correct. Mattthews correlation coefficient is used as the evaluation metric.

\paragraph{SST-2} Stanford Sentiment Treebank \cite{socher2013recursive} consists of 67k
human-annotated movie reviews. The goal is to predict whether each review is positive or negative. Accuracy is used as the evaluation metric.

\paragraph{MRPC} Microsoft Research Paraphrase Corpus \cite{dolan2005automatically} consists of 3.7k
sentence pairs extracted from online news, and the goal to predict if each pair of sentences is semantically equivalent. F1 score from GLUE server is reported as the metric.

\paragraph{QQP} The Quora Question Pairs\footnote{https://data.quora.com/First-Quora-Dataset-Release-Question-Pairs} task consists of 393k question pairs from Quora webiste. The task is to predict whether a pair of questions is semantically equivalent. Accuracy is used as the evaluation metric.

\paragraph{NLI} Multi-Genre Natural Language Inference Corpus \textbf{(MNLI)} \cite{williams2017broad}, 
Question-answering NLI \textbf{(QNLI)} \cite{rajpurkar2016squad} and Recognizing Textual Entailment \textbf{(RTE)}\footnote{Collections of series of annual textual entailment challenges.} are all natural language inference (NLI) tasks, which consist of 393k/108k/2.5k pairs of premise and hypothesis. The goal is to predict if the premise entails the hypothesis, or contradicts it, or neither. Accuracy is used as the evaluation metric. Besides, MNLI test set is further divided into two splits: matched (MNLI-m, in-domain) and mismatched (MNLI-mm, cross-domain), accuracy for both are reported.

\subsection{Implementation Details}
\label{sec:impl_detail}
We mostly follow the pre-training setting from \citet{liu2019roberta}, and use the fairseq implementation \cite{ott2019fairseq}. Specifically, we truncate raw text into sentences with maximum length of 512 tokens, and randomly mask 15\% of tokens as \texttt{[MASK]}. For model architecture, we use a randomly initialized 6-layer Transformer model as the student, and RoBERTa-base with 12-layer Transformer as the teacher. The student model was first trained by using Adam optimizer with learning rate 0.0007 and batch size 8192 for 35,000 steps. For computational efficiency, this model serves as the initialization for the second-stage pre-training with the teacher.
Then, the student model is further trained for another 10,000 steps with KD and the proposed contrastive objective, with learning rate set to 0.0001. We denote this model as CoDIR-Pre.
For ablation purposes, we also train two baseline models with only MLM loss or KD loss, using the same learning rate and number of steps. Similarly, these two models are denoted as MLM-Pre and KD-Pre, respectively. For other hyper-parameters, we use $\alpha_1 = \alpha_2 = 0.1$ for both $\mathcal{L}_{\text{KD}}$ and $\mathcal{L}_{\text{CRD}}$. 

Due to high computational cost for pre-training, all the hyper-parameters are set empirically without tuning.
As there exist many combinations of pre-training loss (MLM, KD, and CRD) and finetuning strategies (standard finetuning with cross-entropy loss, and finetuning with additional CRD loss), a grid search of all the hyper-parameters is infeasible. Thus, for standard finetuning, we search learning rate from \{1e-5, 2e-5\} and batch size from \{16, 32\}. The combination with the highest score on dev set is reported for ablation studies, and is kept {\it fixed} for future experiments. We then fix the hyper-parameters in KD as $\rho=2, \alpha_1=0.7$, and search weight of the CRD loss $\alpha_2$ from \{0.1, 0.5, 1\}, and the number of negative samples from \{100, 500, 1000\}. Results with the highest dev scores were submitted to the official GLUE server to obtain the final results. For fair comparison with other baseline methods, all the results are based on single-model performance.

\begin{table*}[t!]
\centering
\begin{tabular}{|c c c c c c c c|} 
\hline
\multirow{2}{*}{Pre-training Loss} &   \multirow{2}{*}{Finetuning Method}  &         CoLA & SST-2 & MRPC & QNLI & RTE & \multirow{2}{*}{Ave.}\\ 
 [0.5pt] & &(8.5k) & (67k) & (3.7k) & (108k) & (2.5k) & \\
\hline\hline
 $\mathcal{L}_{\text{MLM}}$  &  Standard & 55.4	& 92.0	& 88.0	& 90.2  & 67.1	& 78.5 \\
 $\mathcal{L}_{\text{MLM}}$  &  KD & 57.8    & 92.4  & 89.5  & 90.3  & 68.2  & 79.4\\
 $\mathcal{L}_{\text{MLM}}$  &  CoDIR-Fine & 59.3	& 92.7	& 90.0 & 90.7	& 70.8	& 80.7 \\
 \hline
$\mathcal{L}_{\text{MLM}}$  &  Standard & 55.4	& 92.0	& 88.0	& 90.2   & 67.1	& 78.5 \\
$\mathcal{L}_{\text{MLM}}+\mathcal{L}_{\text{KD}}$  &  Standard & 56.6  & 92.7 & 89.0  & 90.5  & 69.0 & 79.6\\
CoDIR-Pre &  Standard & 57.6  & 92.8 & 90.4 & 90.8 & 71.5 &  80.6\\
 \hline
\end{tabular}
\caption{Ablation study on different combination of pre-trained models and finetuning approach. The results are based on GLUE dev set.}
\label{tab_glue_ablation}
\end{table*}

\begin{table*}[t!]
\centering
\resizebox{1.0\textwidth}{!}{
\begin{tabular}{|c c c c c c|} 
\hline
 Model              &\#Trm Layers    &\#Params   &\#Params (Emb Layer)   &Inference Time (ms/seq)     & Speed-up\\ 
 \hline\hline
 BERT-base      &12              & 109.5M       & 23.8M              & 2.60                      & 1.00$\times$ \\
 RoBERTa-base   &12              & 125.2M       & 39.0M              & 2.53                      & 1.03$\times$\\
 DistilBERT         &6               & 67.0M        & 23.8M              & 1.27                      & 2.05$\times$\\
 CoDIR-Pre        &6               & 82.7M        & 39.0M              & 1.25                      & 2.08$\times$\\ 
 \hline
\end{tabular}
}
\caption{Inference speed comparison between teacher and students. Inference time is measured on MNLI dev set. Speed up is measured against BERT-base, which is the teacher model for other baseline methods. } 
\label{tab_inf_speed}
\end{table*}

\subsection{Experimental Results}
\label{sec:exp_res}
Results of different methods from the official GLUE server are summarized in Table \ref{tab_glue_test}. For simplicity, we denote our baseline approach without using any teacher supervision as ``MLM-Pre + Fine'': pre-trained by using MLM loss first, then finetuning using standard cross-entropy loss. Our baseline already achieves high average score across 8 tasks, and outperforms task-specific model compression methods (such as SID~\cite{aguilar2019knowledge} and BERT-PKD~\cite{sun2019patient}) as well as DistilBERT~\cite{sanh2019distilbert} by a large margin. 

After adding contrastive loss at the finetuning stage (denoted as CoDIR-Fine), the model outperforms the state-of-the-art compression method, TinyBERT with 6-layer Transformer, on average GLUE score. Especially on datasets with fewer data samples, such as CoLA and MRPC, the improved margin is large (+2.5\% and +2.1\%, respectively). Compared to our MLM-Pre + Fine baseline, CoDIR-Fine achieves significant performance gain on almost all tasks (+1.2\% absolute improvement on average score), demonstrating the effectiveness of the proposed approach. The only exception is QQP (-0.1\%) with more than 360k training examples. In such case, standard finetuning may already bring in enough performance boost with this large-scale labeled dataset, 
and the gap between the teacher and student networks is already small (89.6 vs 89.2).

We further test the effectiveness of CoDIR for pre-training (CoDIR-Pre), by applying standard finetuning on model pre-trained with additional contrastive loss
. Again, compared to the MLM-Pre + Fine baseline, this improves the model performance on almost all the tasks (except QQP), with a significant lift on the average score (+1.4\%). We notice that this model performs similarly to the contrastive-finetuning only approach (CoDIR-Fine) on almost all tasks. However, CoDIR-Pre is preferred because it utilizes the teacher's knowledge in the pre-training stage, thus no task-specific teacher is needed for finetuning downstream tasks. Finally, we experiment with the combination of CoDIR-Pre and CoDIR-Fine, and our observation is that adding constrastive loss for finetuning is not bringing in much improvement after already using constrastive loss in pre-training. Our hypothesis is that the model's ability to identify negative examples is already well learned during pre-training.

\begin{table*}[t!]
\centering
\resizebox{1.0\textwidth}{!}{
\begin{tabular}{|c c c c c c c c c|} 
\hline
 Model &        CoLA & SST-2 & MRPC & QQP & MNLI-m/-mm & QNLI & RTE & Ave.\\ 
 [0.5pt] & (8.5k) & (67k) & (3.7k) & (364k) & (393k) & (108k) & (2.5k) & \\
\hline\hline
CoDIR-Fine (\texttt{[CLS]})            & 57.6 & 92.9 & 89.2 & 91.3 & 84.0/84.0 & 90.8 & 70.0 & 82.5\\
CoDIR-Fine  (Mean Pool)     & 59.3	& 92.7 & 90.0    & 91.3    & 84.2/84.2  & 90.7	& 70.8  & \textbf{83.0}\\
 \hline
CoDIR-Fine  (100-neg)      & 57.3 & 92.1 & 88.2 & 91.3 & 84.0/84.0 & 90.4 & 69.3 & 82.1\\
CoDIR-Fine  (500-neg)      & 58.2	& 92.5 & 89.7    & 91.2    & 84.0/84.0  & 90.6	& 70.8  & 82.6\\
CoDIR-Fine  (1000-neg)     & 59.3	& 92.7 & 90.0    & 91.3    & 84.2/84.2  & 90.7	& 70.4  & \textbf{83.0}\\
\hline
\end{tabular}
}
\caption{Ablation study on the use of \texttt{[CLS]} and Mean-Pooling as sentence embedding (upper part) and effect of number of negative examples (neg) for CoDIR-Fine (bottom part). The results are based on GLUE dev set. } 
\label{tab_cls_vs_mean}
\end{table*}

\begin{table*}[t!]
\centering
\begin{tabular}{|c c c c c c c c|} 
\hline
 \multirow{2}{*}{Model} &        CoLA & SST-2 & MRPC & QQP & MNLI-m & QNLI & RTE  \\ 
 [0.5pt] & (8.5k) & (67k) & (3.7k) & (364k) & (393k) & (108k) & (2.5k)  \\
\hline\hline
 Median           & 56.4	& 92.4	& 87.9	& 91.2	& 83.9	& 90.7	& 66.3 \\
 Maximum          & 57.8    & 93.0   & 90.3  & 91.3   & 84.2   & 91.0   & 70.2 \\       
 Standard Deviation & 1.46  & 0.28   & 1.66  & 0.06   & 0.43   & 0.18   & 1.41 \\
\hline
\end{tabular}
\caption{Analysis of model variance on GLUE dev set. Statistical results (median, maximum, and standard deviation) are based on 8 runs with the same hyper-parameters.} 
\label{tab:model_variance}
\end{table*}

\paragraph{Inference Speed}
We compare the inference speed of the proposed CoDIR with the teacher network and other baselines. Statistics of Transformer layers and parameters are presented in Table \ref{tab_inf_speed}. The statistics for BERT$_6$-PKD and TinyBERT$_6$ are omitted as they share the same model architecture as DistilBERT. To test the inference speed, we ran each algorithm on MNLI dev set for 3 times, with batch size 32 and maximum sequence length 128 under the same hardware configuration. The average running time with 3 different random seeds is reported as the final inference speed. Though our RoBERTa teacher has almost 16 million more parameters, it shares almost the same inference speed as BERT-base, because its computational cost mainly comes from the embedding layer with 50k vocabulary size that does not affect inference speed.  By reducing the number of Transformer layers to 6, our proposed student model achieves 2 times speed up compared to the teacher, and achieves state-of-the-art performance among all models with similar inference time.

\subsection{Ablation Studies}

\paragraph{Sentence Embedding}
We also conduct experiments to evaluate the effectiveness of using different sentence embedding strategies. More detailed, based on the same model pre-trained on $\mathcal{L}_{\text{MLM}}$ alone, we run finetuning experiments with contrastive loss on the GLUE dataset by using: $(i)$ \texttt{[CLS]} as sentence embedding; and $(ii)$ mean-pooling as sentence embedding. The results on GLUE dev set are presented in top rows of Table \ref{tab_cls_vs_mean}, showing that mean-pooling yields better results than \texttt{[CLS]} (83.0 vs. 82.5 on average). As a result, we use mean pooling as our chosen sentence embedding for all our experiments.

\paragraph{Negative Examples}
As we mentioned in Section \ref{sec:impl_detail}, the experiments are conducted using 100, 500 and 1000 negative examples. We then evaluate the effect of number of negative examples by comparing their results on GLUE dev set, and the results are presented in the bottom part of Table \ref{tab_cls_vs_mean}. Obviously, for most dataset the accuracy increases as a larger number of negative examples are used during training. Similar observations were also reported in \citet{tian2019contrastive}, and a theoretical analysis is provided in \citet{arora2019theoretical}. 
The only two exceptions are QQP and RTE. As discussed in Section \ref{sec:exp_res}, our CoDIR method seems also not work well on QQP due to the small gap between teacher and student. As for RTE, due to the small number of training examples, the results are quite volatile, which may make the results inconsistent. Besides, the number of negative examples is close to the number of examples per class (1.25k) for RTE, which can also result in the contrastive loss close to 0.

\paragraph{Contrastive Loss}
We first evaluate the effectiveness of the proposed CRD loss for finetuning on a subset of GLUE dev set, using the following settings: ($i$) finetuning with cross-entropy loss only; ($ii$) finetuning with additional KD loss; and ($iii$) finetuning with additional KD loss and CRD loss. Results in Table \ref{tab_glue_ablation} (upper part) show that using KD improves over standard finetuning by 0.9\% on average, and using CRD loss further improves another 1.0\%, demonstrating the advantage of using contrastive learning for finetuning.

To further validate performance improvement of using contrastive loss on pre-training, we apply standard finetuning to three different pre-trained models: ($i$) model pre-trained by $\mathcal{L}_{\text{MLM}}$ (MLM-Pre); ($ii$) model pre-trained by $\mathcal{L}_{\text{MLM}} + \mathcal{L}_{\text{KD}}$ (KD-Pre); and ($iii$) model pre-trained by $\mathcal{L}_{\text{MLM}} + \mathcal{L}_{\text{KD}}+\mathcal{L}_{\text{CRD}}$ (CoDIR-Pre). Results are summarized in Table \ref{tab_glue_ablation} (bottom part). Similar trend can be observed that the model pre-trained with additional CRD loss performs the best, outperforming MLM-Pre and KD-Pre by 1.9\% and 1.0\% on average, respectively.

\paragraph{Model Variance}
Since different random seeds can exhibit different generalization behaviors, especially for tasks with a small training set (\emph{e.g.,} CoLA ), we examine the median, maximum and standard deviation of model performance on the dev set of each GLUE task, and present the results in Table~\ref{tab:model_variance}. As expected, the models are more stable on larger datasets (SST-2, QQP, MNLI, and QNLI), where all standard deviations are lower than 0.5. However, the model is sensitive to the random seeds on smaller datasets (CoLA, MRPC, and RTE) with the standard deviation around 1.5. These analysis results provide potential references for future work on language model compression. 

\section{Conclusion}
In this paper, we present CoDIR, a novel approach to large-scale language model compression via the use of contrastive loss. CoDIR utilizes information from both teacher's output layer and its intermediate layers for student model training. Extensive experiments demonstrate that CoDIR is highly effective in both finetuning and pre-training stages, and achieves state-of-the-art performance on GLUE benchmark compared to existing models with a similar size. All existing work either use BERT-base or RoBERTa-base as teacher. For future work, we plan to investigate the use of a more powerful language model, such as Megatron-LM \cite{shoeybi2019megatron}, as the teacher; and different strategies for choosing hard negatives to further boost the performance.

\bibliography{emnlp2020}
\bibliographystyle{acl_natbib}

\end{document}